\documentclass[runningheads]{llncs}

\usepackage{graphicx}
\usepackage{graphics}
\usepackage{blindtext}
\usepackage[justification=centering]{caption}
\usepackage[sectionbib,numbers,sort&compress]{natbib}
\usepackage{subcaption}

\usepackage[]{hyperref}

\usepackage{listings}
\usepackage{pgfplots}
\usetikzlibrary{shapes, arrows, calc, positioning}

\usepackage{comment}
\pgfplotsset{compat=1.16}

\makeatletter
\def\thickhline{%
  \noalign{\ifnum0=`}\fi\hrule \@height \thickarrayrulewidth \futurelet
   \reserved@a\@xthickhline}
\def\@xthickhline{\ifx\reserved@a\thickhline
               \vskip\doublerulesep
               \vskip-\thickarrayrulewidth
             \fi
      \ifnum0=`{\fi}}
\makeatother
\newlength{\thickarrayrulewidth}
\makeatletter
\def\thinhline{%
  \noalign{\ifnum0=`}\fi\hrule \@height \thinarrayrulewidth \futurelet
   \reserved@a\@xthinhline}
\def\@xthinhline{\ifx\reserved@a\thinhline
               \vskip\doublerulesep
               \vskip-\thinarrayrulewidth
             \fi
      \ifnum0=`{\fi}}
\makeatother
\newlength{\thinarrayrulewidth}
\makeatletter
\def\thanks#1{\protected@xdef\@thanks{\@thanks
        \protect\footnotetext{#1}}}
\makeatother

\begin{document}


%

\title{Using Decomposed Prompting to Answer Questions on a Course Discussion Board \thanks{Supported by Vector Institute, NSERC, Fujitsu, Amazon Research Award, and the CIFAR AI Chairs Program}}
%
\titlerunning{Decomposed Prompting to Answer Questions}
%
\author{Brandon Jaipersaud\inst{3}\orcidID{0009-0007-0478-7356} \and
Paul Zhang \inst{2}\orcidID{0009-0008-7674-6657} \and 
Jimmy Ba \inst{1,3}\orcidID{0009-0000-9062-4180} \and 
Andrew Petersen \inst{2}\orcidID{0000-0003-1337-7985}  \and 
Lisa Zhang \inst{2}\orcidID{0000-0002-7302-6530}  \and 
Michael R. Zhang \inst{1,3}\orcidID{0009-0000-8281-0856}
}
\authorrunning{B. Jaipersaud et al.}

%

\institute{University of Toronto, Toronto ON, Canada \\
\email{\{jba,michael\}@cs.toronto.edu}\\
\and
University of Toronto Mississauga, Mississauga ON, Canada\\
\email{\{pol.zhang,andrew.petersen,lc.zhang\}@utoronto.ca} \\  
\and
Vector Institute, Toronto ON, Canada\\
\email{brandon.jaipersaud@mail.utoronto.ca}}


%
\maketitle              
%



\begin{abstract}
We propose and evaluate a question-answering system that uses decomposed prompting to classify and answer student questions on a course discussion board. Our system uses a large language model (LLM) to classify questions into one of four types: \textit{conceptual}, \textit{homework}, \textit{logistics}, and \textit{not answerable}. This enables us to employ a different strategy for answering questions that fall under different types. Using a variant of GPT-3, we achieve 81\% classification accuracy. We discuss our system's performance on answering conceptual questions from a machine learning course and various failure modes.




\keywords{Course Discussion Board \and GPT-3 \and Large-Language Models \and Mixture of Experts \and Prompting}
\end{abstract}

\section{Introduction}

Course discussion boards are an important avenue for students to ask course-related questions. However, instructors and teaching assistants spend much time and effort responding to discussion board questions, especially for courses with large enrollment \cite{zylich2020exploring, goel2018jill}. With recent advances in natural language processing, there is an opportunity to leverage general-purpose large language models (LLMs) to assist in answering these questions. While GPT-3 has been used in education to adapt assignments~\cite{wang2022towards, sarsa2022automatic} and explanations~\cite{drori2022neural}, we are not aware of work using LLMs for course Q\&A. Furthermore, non-LLM automated methods for answering student questions have been developed in prior work \cite{zylich2020exploring, feng2006intelligent, goel2018jill}. However, the cost of answering a question incorrectly is high: providing an incorrect answer can be detrimental for students and \textit{increase} course staff workload. 

To mitigate this risk, we propose a LLM-based question-answering system that uses decomposed prompting \cite{decomposed_prompting} to answer student questions. We first prompt a LLM to classify a student question into one of four types: \textit{conceptual}, \textit{homework}, \textit{logistics}, and \textit{not answerable}. Then, depending on the question type, we either ignore the question, or we further prompt a LLM to answer the question.

This Mixture of Experts approach can mitigate risk in a few ways. First, the risk and failure modes of an incorrect answer differ depending on the question type (e.g. a question about a deadline vs. explaining subtle concepts).  Since each expert model can be deployed independently, educators have more control over the system. 
Second, we demonstrate that while LLMs can effectively differentiate between types of questions, different answering approaches and contextual cues may be effective for each question type.
This modular approach suggests a framework for managing risk in real deployments and
allows future work to focus on answering specific types of questions effectively.

Section~\ref{sec:moe} describes our approach.
In Section~\ref{sec:question classification}, we show that \textit{our classification system achieves an accuracy of 81\%}, and analyze how the number of few-shot examples, the task description, and the question type labels affect classification accuracy. In Section 4.2, we use various metrics to show that our conceptual prompt works best for answering conceptual questions from an Intro to ML course. We use human evaluation to validate our automatic metrics and discuss the failure modes of our system on conceptual questions.

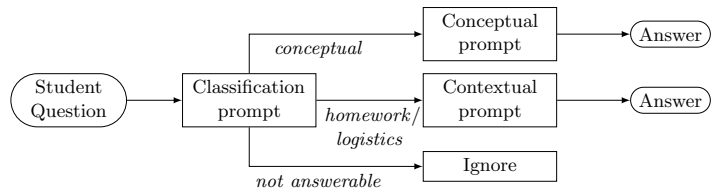
\begin{figure}[t]
    \centering
    \scalebox{0.8}{
    \begin{tikzpicture}[scale=0.7, node distance=3cm]
        \node[draw, rounded rectangle, text width=1.4cm, align=center] (question) {Student Question};
        
        \node[draw, text width=2cm, align=center, right of=question] (classification) {Classification prompt};

        \node[draw, text width=2cm, align=center, right of=classification, xshift=1cm, yshift=1.1cm] (conceptual) {Conceptual prompt};
        \node[draw, text width=2cm, align=center, right of=classification, xshift=1cm, yshift=0cm] (homework) {Contextual prompt};
        \node[draw, text width=2cm, align=center, right of=classification, xshift=1cm, yshift=-1.1cm] (ignore) {Ignore};

        \node[draw, rounded rectangle, right of=conceptual] (conceptual_answer) {Answer};
        \node[draw, rounded rectangle, right of=homework] (homework_answer) {Answer};

        \draw[-latex] (question) -- (classification);
        \draw[-latex] (classification) |- (conceptual) 
            node[pos=0.7, below]{\textit{conceptual}};
        \draw[-latex] (classification) -- (homework) 
            node[midway, below, text width=1.5cm, align=center]{\textit{homework}/ \\ \textit{logistics}};
        \draw[-latex] (classification) |- (ignore) 
            node[pos=0.7, below]{\textit{not answerable}};
        \draw[-latex] (conceptual) -- (conceptual_answer);
        \draw[-latex] (homework) -- (homework_answer);
    \end{tikzpicture}}
    \caption{Mixture of Experts for answering questions using decomposed prompting.}
    \label{fig:moe}
\end{figure}

\section{Mixture of Experts with Decomposed Prompting} \label{sec:moe}
Decomposed prompting \cite{decomposed_prompting} is an approach that uses LLM prompting to decompose a complex problem into simpler subtasks where each subtask can be solved by further prompting a LLM. Our mixture of experts design (Figure~\ref{fig:moe}) decomposes the task of answering student questions into two subtasks: \textbf{Question Classification} and \textbf{Question Answering}. We first classify questions using the prompt in Figure~\ref{fig:prompts} (left). Based on the classification, we further prompt a LLM to answer the question using Figure~\ref{fig:prompts} (right) or ignore the question. 




The \textbf{Question Classification} task classifies student questions by type. As in prior work~\cite{zylich2020exploring}, our question types are based on the course-specific context required to answer the question. 
The four question types we use are:

\begin{enumerate}
    \item \textbf{Conceptual Questions} that can be answered without course-specific context, e.g., \textit{How do we choose the learning rate?}
    \item \textbf{Homework Questions} that require the corresponding homework instructions to be answerable, e.g., \textit{What does z refer to in Lab 1?}.
    \item \textbf{Logistics Questions} that require the course syllabus to be answerable, e.g., \textit{Which room is the midterm in?}
    \item \textbf{Not Answerable Questions} that require human intervention, e.g.,
    \textit{The instructor isn't here. Have office hours been cancelled?}

\end{enumerate}

The \textbf{Question Answering} task proposes answers to a question. Our intention is for each question type to have a different prompting strategy: for example, \textit{conceptual} questions do not require specialized context in the prompt. However, \textit{homework} questions require that we identify and provide relevant sections of the assignment handout. Likewise, \textit{logistics} questions require context from the course syllabus. Finally, \textit{not answerable} questions are left to the instructors. 




\section{Results and Analysis}

\subsection{Question Classification}\label{sec:question classification}

We evaluate our classification system on 72 historical student questions randomly sampled from the Fall 2022 instance of an upper-level Intro to ML course at a research-intensive institution in North America. 
We choose upper-year courses since, in our experience, there are more conceptual questions in these courses. 
To form the ground-truth type labels for the 72 questions, three course staff manually classified each question into the four types described in Section~\ref{sec:moe}. 27 of the 72 annotated questions had a disagreement between two of the annotators, and we take the majority label as the ground truth. For 3 questions, all annotators disagreed on the ground truth label; we discard these questions from our analysis. Of the remaining 69 questions with ground truth labels, there are 13 \textit{conceptual}, 34 \textit{homework}, 8 \textit{logistics}, and 14 \textit{not answerable} questions. 






The LLM we use for this task is \textit{text-davinci-003}, an InstructGPT~\cite{ouyang2022training} variant of OpenAI's GPT-3 model fine-tuned on human instructions. Our prompt, which includes 31 in-context examples, is shown in Figure~\ref{fig:prompts} (left); we justify our choice of prompt below. In Table~\ref{table:classification f-score}, we give a breakdown of the F-score by type; the total classification accuracy of our system is 81\%.





\begin{table}[t]
\centering
\caption{Precision, recall, and F-score for question classification type. The overall accuracy is 81\%.}\label{table:classification f-score}
\begin{tabular}{l|c|c|c|c|c}
\thinhline
\textbf{Type} & \textbf{Count} & \textbf{\#Correct} & \textbf{Precision} & \textbf{Recall} &\textbf{F-Score}\\
\thickhline
\textit{conceptual} & 13 & 11 & 0.79 & 0.85 & 0.81 \\
\thinhline
\textit{homework} &   34 & 27 & 0.96 & 0.79 & 0.87  \\
\thinhline
\textit{logistics} &  8 & 5 & 0.63 & 0.63 & 0.63 \\
\thinhline
\textit{not answerable} & 14& 13 & 0.68 & 0.93 & 0.79 \\
\thinhline
\end{tabular}
\end{table}

\textbf{Task Description} Our prompt begins with a task description outlining when a question belongs to each type (Figure~\ref{fig:prompts} (left)). To test the importance of this description, we evaluate three alternatives. With our descriptive prompt, the classification accuracy is 81\%. Using no task description reduces the accuracy to 74\%. Using only the first sentence of the description gives the lowest performance at 72\%. Adding a sentence stating \textit{Questions that point out corrections or typos should be classified as ``homework"} gives an accuracy of 77\%.



\textbf{Number of Few-shot  Examples} 
We use few-shot prompting and provide example questions and classifications (Figure~\ref{fig:prompts} (left)).
To test the importance of these in-context examples, we vary the number of examples in our prompt. Table~\ref{table:classification sensitivity} shows that using 31 examples produces the highest classification accuracy.

\begin{table}[t]
\centering
\caption{Classification accuracy by number of few-shot examples.}\label{table:classification sensitivity}
\begin{tabular}{l|r|r|r|r|r|r|r|r}
\hline
\textbf{Examples} & 0 & 2 & 4 & 8 & 16 & 24 & 31 & 42 \\
\hline
\textbf{Accuracy} & 42\% & 61\% & 58\% & 67\% & 65\% & 77\% & 81\% & 70\% \\
\hline
\end{tabular}
\end{table}

\textbf{Question Type Labels} The classification performance is sensitive to the choice of type labels used in the prompt. Renaming the types to \textit{a, b, c, d} reduces accuracy to 70\%. Renaming the types to \textit{directly answerable}, \textit{needs course material}, \textit{needs administrative material}, \textit{not answerable} reduces accuracy to 74\%. Renaming the types to \textit{conceptual}, \textit{needs course material}, \textit{needs administrative material}, \textit{not answerable} reduces accuracy to 75\%.

\subsection{Question-Answering System} \label{sec:qa_system}

In this section, we present an initial effort to answer \textit{conceptual} questions.
In addition to the 69 questions from earlier, we include 63 additional questions from the Winter 2023 offering of the same Intro to ML course. 
Of these 63 questions, there are 20 \textit{conceptual}, 31 \textit{homework}, 8 \textit{logistics} and 4 \textit{not answerable} questions. 


As before, we use \textit{text-davinci-003}, now with the conceptual prompt shown in Figure~\ref{fig:prompts} (right). We use a temperature of 0.7 for answer generation since we have observed that this temperature produces more descriptive answers.









\textbf{Difficulty of each Question Type} To justify focusing our attention on conceptual questions, 
we start by generating model answers to all 132 questions regardless of question type, using the conceptual prompt in Figure~\ref{fig:prompts} (right). We then evaluate the model's answers using the following metrics:
\begin{itemize}
    \item The \textit{Cosine similarity} between the embeddings of the model answer and the instructor answer, generated with Cohere's Embedding API.
    \item The \textit{ROUGE score} between model and instructor answer to measure textual similarity between the answers.
    \item The \textit{Perplexity of the instructor answer}, which measures how likely the model is to generate the instructor answer.
\end{itemize}




The result of applying these metrics across all 132 questions is shown in Table~\ref{table:autoeval}. The LLM performs best on conceptual questions across all three metrics.
This agrees with our intuition that conceptual questions are easier for a general-purpose LLM to answer since they do not require course-specific details.



\begin{table}[t]
\centering
\caption{Similarity between instructor and model answer by question type.}\label{table:autoeval}
\begin{tabular}{l|c|c|c|c}
\thinhline
\textbf{Question Type} & \textbf{Count} & \textbf{Cosine Similarity} & \textbf{ROUGE1/2/L} & \textbf{Perplexity}\\
\thickhline
\textit{conceptual} &  34 & 0.62 & 0.30/0.07/0.18 & 7.61 \\
\thinhline
\textit{homework} &   59 & 0.48 & 0.23/0.06/0.16 & 12.73 \\
\thinhline
\textit{logistics} &  16 & 0.43 & 0.17/0.04/0.14 & 13.01 \\
\thinhline
\textit{not answerable} & 23 & 0.52 & 0.19/0.03/0.13 & 34.32 \\
\thinhline
\end{tabular}
\end{table}

\textbf{Human Evaluation on Conceptual Questions}
The Intro to ML instructors labeled the 28 model answers to conceptual questions with \textit{good answer} or \textit{bad answer}, depending on whether the answer correctly and appropriately addresses the student's question. Table~\ref{table:humanfeedback} shows the result of applying the automated metrics to each feedback category. The \textit{good} answers are semantically and textually closer to their corresponding instructor answers. This validates our use of the metrics in Table~\ref{table:autoeval}.

Furthermore, we see that 8/28 (29\%) of annotated answers were labeled with \textit{good answer}. Of the 20 bad answers, we observe the following distribution of failure modes:
  3 answers address questions that are misclassified as \textit{conceptual} (rather than \textit{homework}).
  5 answers are factually incorrect. 
  For instance: \textit{``The training set can be used to tune hyperparameters"}. 
  3 answers are correct but inappropriate for the student's level of knowledge. 
  For example: \textit{``Gradient descent can be used to optimize hyperparameters, although this is less common"} when this form of optimization was not discussed in the course.
  The remaining bad answers suffer from issues such as misunderstanding the question, incoherence, making incorrect assumptions, or adding irrelevant information to the answer. 
  





%

\begin{table}[t]
\centering
\caption{Human and automatic evaluation of conceptual answers.} \label{table:humanfeedback}
\begin{tabular}{l|c|c|c|c}
\thinhline
\textbf{Feedback} & \textbf{Count} & \textbf{Cosine Similarity} & \textbf{ROUGE1/2/L} & \textbf{Perplexity}\\
\thickhline
\textit{good answer} & 8 & 0.83 & 
 0.65/0.51/0.57 & 3.05\\
\thinhline
\textit{bad answer} & 20 & 0.61 & 0.31/0.07/0.18  & 6.35\\
\thinhline
\end{tabular}
\end{table}

\begin{figure}[t]
     \centering
     \begin{subfigure}[b]{0.8\textwidth}
         \centering
\begin{lstlisting}[
    basicstyle=\tiny\ttfamily
]
Task Description: Classify each question posted on an undergraduate course discussion board into one of the following 4 types: conceptual", "homework", "logistics" or "not answerable".
A question that requires instructor intervention should be classified as "not answerable". Questions that point out contradictions in assignment instructions and deadlines should be classified as "not answerable". Conceptual questions should be classified as "conceptual". Homework and lab questions that provide enough information to be answered should be classified as "conceptual". Questions that need course content related context to be answered such as an assignment handout should be classified as "homework". Questions that need logistical context to be answered such as a course syllabus should be classified as "logistics". 

Question: Is the A1 Q2 code for debugging a neural network correct? It says we should debug using a large dataset. However, using a small dataset seems to make more sense here.
Classification: homework

... (30 more question/classification pairs)

Question: <student question>
Classification:
\end{lstlisting}
     \end{subfigure}
     \hfill
     \begin{subfigure}[b]{0.15\textwidth}
         \centering
\begin{lstlisting}[
    basicstyle=\tiny\ttfamily
]
Task: Answer the following question that was posted by a student on the class discussion board for an introductory machine learning course. Your answer should be truthful, concise and helpful to the student. 

Question: <student question>

Answer:
\end{lstlisting}
     \end{subfigure}
     \hfill
        \caption{Prompt used for classification (left) and conceptual Q\&A (right).}
        \label{fig:prompts}
\end{figure}

\newenvironment{prompt}[1]
{\paragraph{\hspace{0.3cm}#1}\vspace{-0.8\baselineskip}
    \begin{quote}\small
    }
    {\end{quote} 
    }



\section{Conclusion}

We demonstrated that decomposed prompting is a useful strategy for classifying and answering student questions due to their context-dependent or unanswerable nature. Furthermore, our results have shown that LLMs can classify student questions with an accuracy of 81\%, but yield poor performance when answering conceptual questions. Many of the poor answers arose due to misalignment with the preferences of a course instructor. Future work could aim to fine-tune the model on course discussion board questions. 
An interesting direction for future work is to combine LLMs with semantic processing or text extraction techniques \cite{goel2018jill, zylich2020exploring} which should enable prompted LLMs to handle homework and logistics questions.

\bibliographystyle{splncs04}
\bibliography{references}

\begin{thebibliography}{8}
\providecommand{\natexlab}[1]{#1}
\providecommand{\url}[1]{\texttt{#1}}
\providecommand{\urlprefix}{URL }
\expandafter\ifx\csname urlstyle\endcsname\relax
  \providecommand{\doi}[1]{doi:\discretionary{}{}{}#1}\else
  \providecommand{\doi}{doi:\discretionary{}{}{}\begingroup \urlstyle{rm}\Url}\fi

\bibitem[{Drori et~al.(2022)Drori, Zhang, Shuttleworth, Tang, Lu, Ke, Liu, Chen, Tran, Cheng et~al.}]{drori2022neural}
Drori, I., Zhang, S., Shuttleworth, R., Tang, L., Lu, A., Ke, E., Liu, K., Chen, L., Tran, S., Cheng, N., et~al.: A neural network solves, explains, and generates university math problems by program synthesis and few-shot learning at human level. Proceedings of the National Academy of Sciences \textbf{119}(32), e2123433119 (2022)

\bibitem[{Feng et~al.(2006)Feng, Shaw, Kim, and Hovy}]{feng2006intelligent}
Feng, D., Shaw, E., Kim, J., Hovy, E.: An intelligent discussion-bot for answering student queries in threaded discussions. In: Proceedings of the 11th international conference on Intelligent user interfaces, pp. 171--177 (2006)

\bibitem[{Goel and Polepeddi(2018)}]{goel2018jill}
Goel, A.K., Polepeddi, L.: Jill watson: A virtual teaching assistant for online education. In: Learning engineering for online education, pp. 120--143, Routledge (2018)

\bibitem[{Khot et~al.(2022)Khot, Trivedi, Finlayson, Fu, Richardson, Clark, and Sabharwal}]{decomposed_prompting}
Khot, T., Trivedi, H., Finlayson, M., Fu, Y., Richardson, K., Clark, P., Sabharwal, A.: Decomposed prompting: A modular approach for solving complex tasks (2022), \doi{10.48550/ARXIV.2210.02406}

\bibitem[{Ouyang et~al.(2022)Ouyang, Wu, Jiang, Almeida, Wainwright, Mishkin, Zhang, Agarwal, Slama, Ray et~al.}]{ouyang2022training}
Ouyang, L., Wu, J., Jiang, X., Almeida, D., Wainwright, C.L., Mishkin, P., Zhang, C., Agarwal, S., Slama, K., Ray, A., et~al.: Training language models to follow instructions with human feedback. arXiv:2203.02155  (2022)

\bibitem[{Sarsa et~al.(2022)Sarsa, Denny, Hellas, and Leinonen}]{sarsa2022automatic}
Sarsa, S., Denny, P., Hellas, A., Leinonen, J.: Automatic generation of programming exercises and code explanations using large language models. In: Proceedings of the 2022 ACM Conference on International Computing Education Research-Volume 1, pp. 27--43 (2022)

\bibitem[{Wang et~al.(2022)Wang, Valdez, Basu~Mallick, and Baraniuk}]{wang2022towards}
Wang, Z., Valdez, J., Basu~Mallick, D., Baraniuk, R.G.: Towards human-like educational question generation with large language models. In: Artificial Intelligence in Education: 23rd International Conference, AIED 2022, Durham, UK, July 27--31, 2022, Proceedings, Part I, pp. 153--166, Springer (2022)

\bibitem[{Zylich et~al.(2020)Zylich, Viola, Toggerson, Al-Hariri, and Lan}]{zylich2020exploring}
Zylich, B., Viola, A., Toggerson, B., Al-Hariri, L., Lan, A.: Exploring automated question answering methods for teaching assistance. In: Artificial Intelligence in Education: 21st International Conference, AIED 2020, Proceedings, Part I 21, pp. 610--622, Springer (2020)

\end{thebibliography}

\end{document}